\documentclass[letterpaper, 10 pt, conference]{ieeeconf} 
\pdfoutput=1
\IEEEoverridecommandlockouts 
\usepackage{graphicx}
\usepackage[utf8]{inputenc}

\title{\LARGE \bf
A GAN-based Approach for Mitigating Inference Attacks in Smart Home Environment}

\author{ \parbox{5 in}{\centering Olakunle Ibitoye, Ashraf Matrawy,  and  Omair Shafiq
         \thanks{}\\
         School of Information Technology\\
         Carleton University, Ottawa, Canada\\
         \tt\small Email: \{Kunle.Ibitoye, Ashraf.Matrawy, Omair.Shafiq, \}@carleton.ca
}
}

\begin{document}

\maketitle
\thispagestyle{empty}
\pagestyle{empty}

\begin{abstract}

The proliferation of smart, connected, always listening devices have introduced significant privacy risks to users in a smart home environment. Beyond the notable risk of eavesdropping, intruders can adopt machine learning techniques to infer sensitive information from audio recordings on these devices, resulting in a new dimension of privacy concerns and attack variables to smart home users. Techniques such as sound masking and microphone jamming have been effectively used to prevent eavesdroppers from listening in to private conversations. In this study, we explore the problem of adversaries spying on smart home users to infer sensitive information with the aid of machine learning techniques. We then analyze the role of randomness in the effectiveness of sound masking for mitigating sensitive information leakage. We propose a Generative Adversarial Network (GAN) based approach for privacy preservation in smart homes which generates random noise to distort the unwanted machine learning-based inference. Our experimental results demonstrate that GANs can be used to generate more effective sound masking noise signals which exhibit more randomness and effectively mitigate deep learning-based inference attacks while preserving the semantics of the audio samples.

Index terms - privacy, IoT, generative adversarial networks, information leakage, smart home

\end{abstract}

\section{Introduction}\label{intro}

The presence of "always listening" devices in a user's environment poses significant privacy concerns, as an adversary may leverage these devices to eavesdrop on a user's private conversations. With the proliferation of IoT smart home devices, the number of microphone bearing devices in residential homes has increased exponentially over the past decade resulting in an increased attack surface for adversaries. A typical case study in \cite{Checkmarx2018} showed that researchers with minimal coding effort converted an Amazon Echo smart speaker into a spy device for eavesdropping on homeowners. Similar studies \cite{lau2018alexa} \cite{zeng2017end} have suggested a wide range of adversaries including IoT platform owners, app developers, device manufacturers and solution providers.

The ability of GANs to create high dimensional data has been researched \cite{de2018pseudo}. In this study, we seek to correlate the relationship between high dimensional data in GAN generated audio samples and increased randomness.

The use of sound masking noise signals has been an important technique in protecting user privacy against eavesdroppers who attempt to gain unauthorized access to users' private conversations. One recommended approach for protecting user privacy in smart homes from always listening devices is the use of sound masking by adding white noise to  audio signals \cite{jiang2017research}. White noise includes all frequencies at equal energy. However, it is more desirable to generate audio signals that sound more comfortable to listeners. As such, only the specific frequency spectrum that are required to increase privacy are produced with resultant minimal distraction.

Providers and vendors of smart home IoT devices have argued that connected, always listening devices such as Amazon Echo speakers implement a temporary buffer \cite{godwin2019future} that prevents the device from continuously recording user conversations. In addition, users can review and delete their voice recordings either through the app or by voice commands. While these techniques are a reasonable proposition, their effectiveness is not yet proven in preserving user privacy, especially since the attack surface increases with Internet connectivity and accessibility to various third-party apps. Moreover, this requires having a complete trust in several providers (hardware, software, etc.) and the insiders within the organization, which may not be guaranteed all the time.

In this study, we explore risks beyond eavesdropping and consider information leakage in smart home environments. An  information  leakage attack  provides  a  larger  attack surface because the adversary can deduce or infer sensitive information with the aid of computation and machine learning techniques. For example, an adversary can infer that there is an infant child in the home, can infer the race and gender of the occupants, or activities being performed in a home by merely running inference attacks on the smart home devices \cite{bugeja2016privacy} \cite{apthorpe2019keeping}.

Sound-based inference attacks  may provide  greater  incentives  to  adversaries. For example, they can  get thousands of users to download an app, and infer certain sensitive information such as behavioral patterns for a large number of people, which can then be used for commercial purposes such as advertising and sales targeting \cite{huang2019iot}. An adversary may also use such information for more malicious purposes, which  could  jeopardise  the safety  of  the  smart  home occupants,  such  as  inferring  home  occupancy  and  planning a robbery attack \cite{bastos2018internet}.

Our contributions from this study are twofold. In our \textbf{first contribution}, we demonstrate that GAN generated noise results in better performance in mitigating machine learning-based information leakage inference in smart home environments due to the increased randomness in the GAN noise. We show the relationship between randomness in audio signals and the effectiveness of a sound-masking noise signal in preventing sensitive information leakage.

For our \textbf{second contribution}, we introduce a novel Generative Adversarial Network (GAN) structure for producing sound-masking noise signals that are proven to be truly random. We adopt existing frameworks for measuring the randomness element in discrete signals and demonstrate that the GAN based audio noise signals have more entropy-based randomness compared to digitally generated white noise signals.

The novelty of our research is demonstrated in the following ways. To the best of our knowledge, this is the first study to investigate the use of GAN-based noise for mitigating sound-based privacy leakage inference attacks targeted against smart home environments.

Also, this is the first study to the best of our knowledge to investigate the effect of randomness in the ability of a sound masking noise signal to mitigate sensitive information leakage. Our findings show that information leakage mitigation is strongly correlated with the randomness element in the sound masking audio signal. 

We further demonstrate that GANs can generate noise signals which can effectively mitigate sound-based privacy inference attacks while maintaining the semantics of the audio signal, as shown in section \ref{sra_results}. 

\section{Related Work}
Existing research work for privacy preservation / information leakage prevention with noise distortion has focused on signal jamming – to distort the signal and prevent an eavesdropper from listening. No existing solution has utilized generative adversarial networks to create the noise distortion. Similar work \cite{chen2020wearable} has also used ultrasonic transmission to jam nearby microphones .

Lei et al. \cite{lei2017insecurity} proposed the use of a physical presence based access control mechanism which ensures that physical presence is detected before activating the "wake word" in voice assistants as a security measure or before accepting voice commands from a voice-activated digital assistant. While this technique is effective, a carefully crafted malware can effectively fool this safeguard \cite{an2018malware}. In addition, the presence of a home occupant is not a deterrent for an intruder who is deploying and executing malware remotely since the malicious app will most likely operate saliently and quietly. The authors in \cite{10.1145/3133956.3133962}  proposed a Doppler radar-based liveliness detector to prevent spoofing attacks on voice assistants and ensure a human is present before accepting voice commands. The work of \cite{gao2018traversing} proposes a framework that implements a solution that jams the device microphone until the user issues a voice command.

Authors in \cite{liu2019ppgan} \cite{frigerio2019differentially} discussed the limitations of GANs in that it learns the data distribution from the dataset and tends to remember the training samples, which can be used to infer sensitive information from the dataset. The authors thus propose approaches to incorporate privacy preservation techniques into the structure of the GAN. 

Researchers have investigated the use of noise audio signals for privacy preservation in smart environments, such as in \cite{asatilla2018information}, where a noise generator was proposed for preserving privacy in smart tactical platforms.  In \cite{tung2019exploiting}, researchers exploited the use of audio masking for preventing sensitive information leakage in smartphones.

The feasibility of inferring sensitive information in smart homes has been studied from numerous contexts. In \cite{copos2016anybody}, user activities such as walking, sleeping etc. could be inferred by observing the network traffic in a smart home. Even when such traffic is encrypted as in \cite{apthorpe2017spying}, an adversary can still perform information leakage attacks on the smart home, compromising user privacy and confidentiality.

Several attempts have been made to explore the usage of GAN's in network security. \cite{usama2019generative} proposed the use of GAN's for defending against adversarial attacks in network security.

From our literature review, we observed that no published work had explored the use of Generative Adversarial Networks (GANs) to generate audio noise signals to mitigate audio inference attacks in smart home environments. Our research, therefore, seeks to close this gap.

\section{Problem Statement and Proposed Solution}
Several users have installed various IoT devices to make their homes smarter. These devices are always connected, measuring, and collecting data about the environment. An adversary can use the information from those sensors to infer sensitive information about the occupants of the home. This raises significant privacy concerns. For example, researchers have been able to infer the TV content of home users by listening to the sound from the TV \cite{xu2014watching}. 

While it is easy for someone who is familiar with the movie to tell just by listening to the audio sound if the person is in close proximity to the home, the proliferation of smart, connected devices that are always listening creates a larger attack surface. This means that IoT devices or smartphones could be accessed remotely without the owners' authorization or consent to deduce and infer such sensitive content. We term this for the scope of this study, as an inference attack.

As machine learning algorithms are becoming more sophisticated, adversaries will utilize machine learning and deep learning techniques to compromise the privacy of smart home users. In one of such attack variations, an adversary could seek to intercept digital voice assistants, which are very common in many smart home environments and are incorporated into various devices, such as smart speakers, smart refrigerators, and smartphones. In order to prevent sensitive information leakage from digital voice assistants, which are heavily integrated into smart home devices, we need to understand how an adversary can achieve such information leakage and the risk associated with it as well as the consequences of such leakage.

\subsection{What is an Inference Attack?}\label{prob}
\begin{figure}[hb!]
	\centering
	\includegraphics[width=0.99\linewidth,keepaspectratio=true]{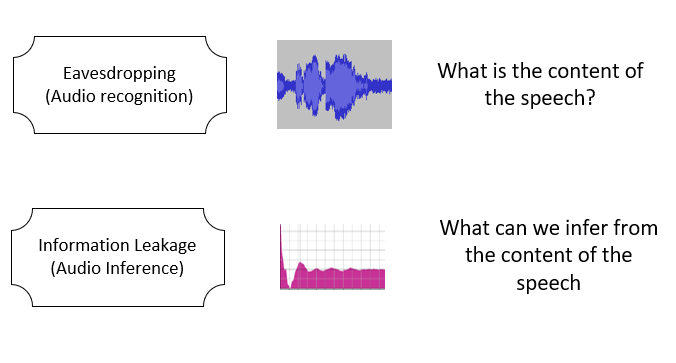}
	\caption{Eavesdropping vs Inference}
	\label{fig:differen}
\end{figure}

In the context of this study, an inference attack occurs when an external party infers sensitive information from data that they have access to \cite{copos2016anybody}. Deep neural networks (DNN) are widely used for various audio processing tasks which fall under two broad categories of audio analysis and audio synthesis/transformation \cite{purwins2019deep}.  Our study represents a borderline between these two categories where we draw a distinction between the two categories and differentiate between audio recognition and audio inference. In this study we focus on audio inference aspects whereby our target is not to recognize what was said, but what could be inferred from what was said. Fig. \ref{fig:differen} illustrates the difference between eavesdropping and inference.

Consider a similar case scenario in which a user downloads a malicious app which exploits the "always listening" capability of a smart device and then runs a script to infer the user's movie preferences. This is also an  example of an information leakage attack.

\subsection{MaskGAN: Our Proposed Solution}\label{maskgan_proposed}
In our proposed solution, we utilize Generative Adversarial Networks (GANs) \cite{goodfellow2014generative} to generate sound masking audio noise to mitigate the information leakage as a result of the machine learning-based inference. More details about the MaskGAN structure is provided in section \ref{maskgan}. The advantage of our proposed solution is twofold. First, GANs due to the lack of a deterministic bias \cite{goodfellow2014generative} can generate synthetic data samples that are truly random. Our objective in this study is to investigate if the noise generated by MaskGAN can mitigate information leakage while preserving the semantics of the audio as shown in the results section \ref{results}. The second advantage is that our solution is independent of the smart home device manufacturer, vendor or solution provider and is completely within the control of the user. We ensure that the noise generated by the MaskGAN does not exceed a sound intensity of 45db which is within the comfort zone for human hearing \cite{fincher2009standards}.

In this study, we seek to understand the role which randomness plays in sound masking privacy preservation. We conduct experiments to determine if our GAN-based approach for generating sound masking noise signals can produce audio noise signals that exhibit more randomness compared to white noise.

\subsection{Research Questions}

Our study seeks to answer the following research questions.

1. Are GAN-generated noise samples effective for information leakage prevention in smart home environments to deter various adversaries from inferring sensitive information from user conversations?

2. Can GAN-generated noise samples be used to deter adversaries from inferring sensitive information from smart home devices, while maintaining the semantics of the audio samples?

3. Are GAN-generated noise samples more random compared to white noise? What role does randomness play in improving the ability of privacy-preserving sound masking techniques to prevent the risk of inferring sensitive information from "always listening" smart home devices?

Our findings to these three research questions are reported in the results section \ref{results}.

\section{Threat Model}\label{threat}

\begin{figure}[h!]
	\centering
	\includegraphics[width=0.99\linewidth,keepaspectratio=true]{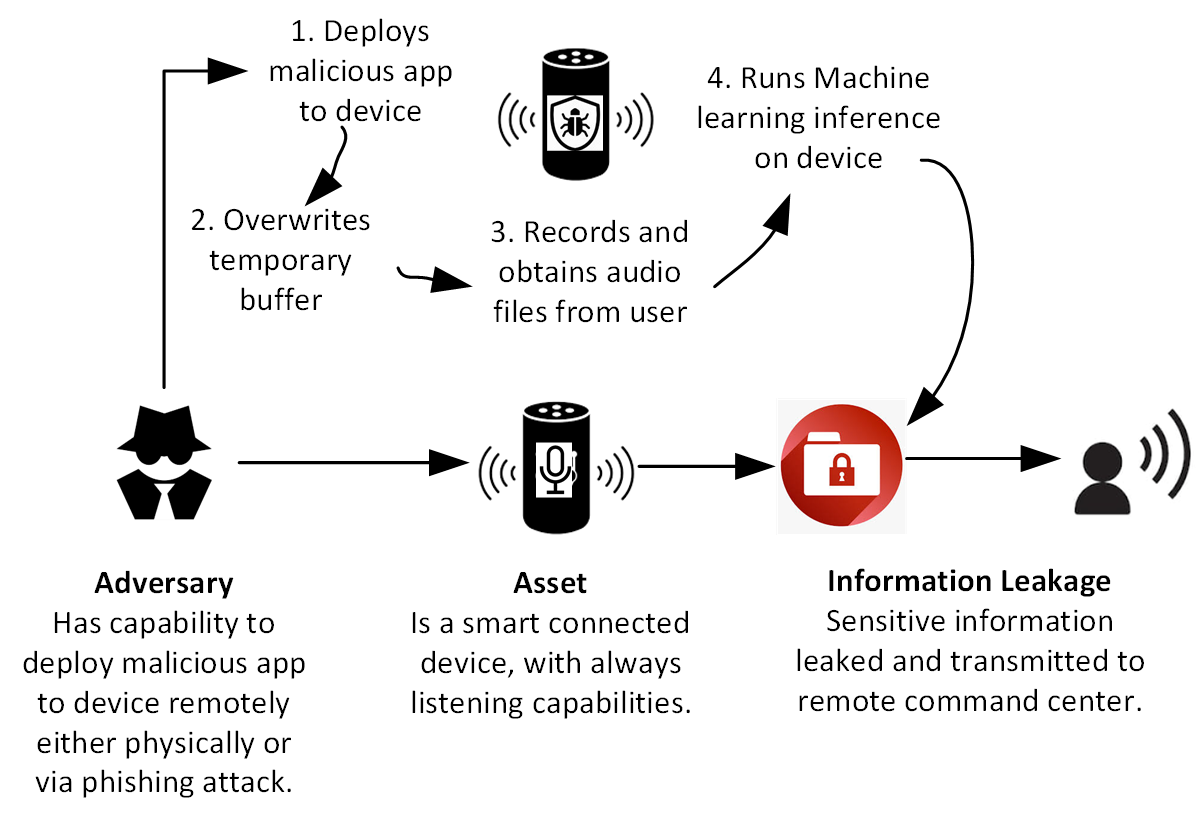}
	\caption{Threat Model}
	\label{fig:threat_model}
\end{figure}

The threat model in our study assumes an information leakage scenario in which an adversary accesses audio files from an "always listening" connected device in a smart home, and infers sensitive information such as user demographics or activities of the home occupants. Fig. \ref{fig:threat_model} illustrates our threat model.

We assume the adversary is anyone other than the legitimate owner of the smart home device's data. The adversary could be a device manufacturer, an insider, an authorized third-party app developer, an unauthorized intruder such as one who deploys a malicious app, or a possible state actor as discussed in \cite{cauley2006nsa}. The adversaries have different capabilities but it is assumed that all adversaries can access the smart home device either physically or remotely. The threat model illustrated in Fig. \ref{fig:threat_model} illustrates an unauthorized third-party adversary who deploys a malicious app onto the smart device either through physical access or through a phishing attack. The malicious app compromises any existing protection e.g., the temporary buffer which prevents the smart device from continuously recording conversations \cite{godwin2019future}.

Different adversaries follow the same pattern of attack against the end-user with the same end goal  - in which sensitive information the smart device user has not given consent to, is inferred and ultimately used for the adversary's gain. We note that there are various capabilities for the  different adversaries, and various types of attacks that can be launched by each of the adversaries based on their capabilities such as eavesdropping, inference attacks or other malicious purposes. For the scope of this study, however, we focus only on inference attacks, in which the adversary applies machine learning techniques to infer sensitive information from the audio recordings from those devices.

It is also assumed that the adversary has full knowledge of the model in what could be referred to as a white box attack. The adversary gains access to the recordings and carries out an inference attack on the recorded conversation. In our experimental approach, three different Deep Neural Network (DNN) models namely the Convolutional Neural Network (CNN), Recurrent Neural Network (RNN) and the Convolutional Recurrent Neural Network (CRNN), were used to infer sensitive information from the audio recordings.


\section{Solution Overview}

Generative adversarial networks (GAN's) \cite{goodfellow2014generative} belong to the set of unsupervised deep learning algorithms known as generative models which learn the underlying hidden structure of given data without specifying a target value. Generative models typically generate synthetic inputs \(x'\), given an input data \(x\), by learning the intrinsic distribution function \(p(x)\) of the input data. In contrast to discriminative models which tend to model the conditional probability distribution function \(p(y|x\), for a given function \(y(x)\),  generative models are direct density implicit models which model \(p(x)\) without attributing the probability distribution function.

\subsection{Audio Features Representation}
Feature representation of the audio signal plays an important role in the deep learning model's ability to infer sensitive information from an audio sample. We consider the task of feature representation for this study different from that of audio classification tasks since the features that serve best for audio classification might not adequately suffice for inferring sensitive information \cite{avci2019pattern}. As a basic foundation, the upper layers of a DNN are best suited for performing feature extraction. In contrast, the lower layers are established to perform class discrimination \cite{mohamed2012understanding} to output the target class. While it is possible to use Mel frequency cepstral coefficients (MFCCs) for the acoustic feature representation, since our study utilizes deep learning models, this approach is ignored because spatial information is lost from the MFCC. 

\begin{figure}[h!]
	\centering
	\includegraphics[width=0.5\linewidth,keepaspectratio=true]{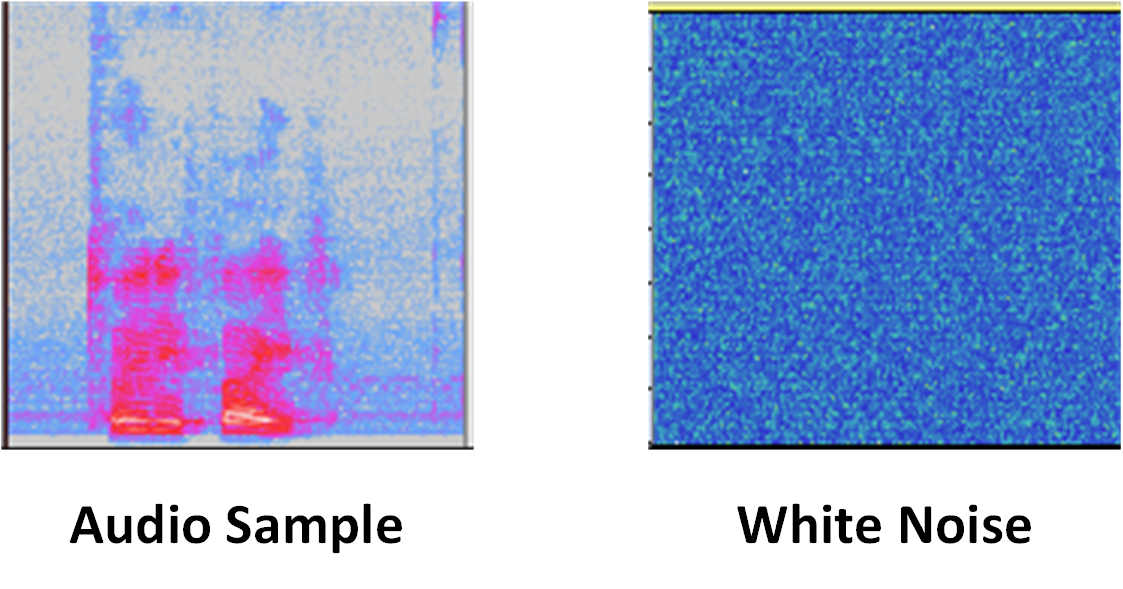}
	\caption{Mel spectrogram of audio sample vs. white noise}
	\label{fig:audio1}
\end{figure}

An alternative representation known as the spectrogram consists of a temporal sequence of spectra. It can be obtained by omitting the Discrete Cosine Transform (DCT) to yield the log-mel spectrum \cite{purwins2019deep}. Fig. \ref{fig:audio1} shows an illustration of spectrogram images for an audio sample in our dataset and a white noise sample that was generated for our experiment.

Even though the spectrograms are similar to images, the approach for audio processing using DNN's is considered to be different from image classification due to the variation in value distribution for audio samples as compared to image samples.

We desist from using the time-domain waveform samples of the audio representation since they do not capture sufficient spatial information which is crucial for our machine learning model and technique.

\subsection{Neural Network Models}
In the past, it was common practice to model and analyze audio signals using Gaussian mixture due to their mathematical elegance \cite{purwins2019deep}. However, in recent times, Deep Neural Networks (DNNs) have shown to be more accurate for audio processing and classification tasks \cite{hinton2012deep}. In this study we examine the performance of three types of neural network models namely - Convolutional Neural Network (CNN), Recurrent Neural Network (RNN), and Convolutional Recurrent Neural Network (CRNN) for our task of inferring sensitive information from audio samples. 

A Convolutional Neural Network (CNN) consists of a series of convolutional layers that are passed through pooling layers, followed by one or more dense layers. Since our study is based on spectral input features from the Mel spectrogram, a 2-dimensional time-frequency convolution is used for computing the feature maps, which are further downsampled by the pooling layers. The optimal parameters for the CNN are obtained experimentally based on the validation error observed during the training process. Recurrent Neural Networks (RNN), are well suited for sequence modeling tasks such as audio processing \cite{lipton2015critical} due to the fact that they intrinsically model the temporal dependency in the input features. A Convolutional Recurrent Neural Network (CRNN) \cite{choi2017convolutional} is an extension of the CNN in which an RNN is implemented to process the output of a CNN. While the purpose of the convolutional layers is to perform feature extraction, the recurrent layers enable the model to make sense of the longer temporal context. 

\begin{figure}[h!]
	\centering
	\includegraphics[width=0.9\linewidth,keepaspectratio=true]{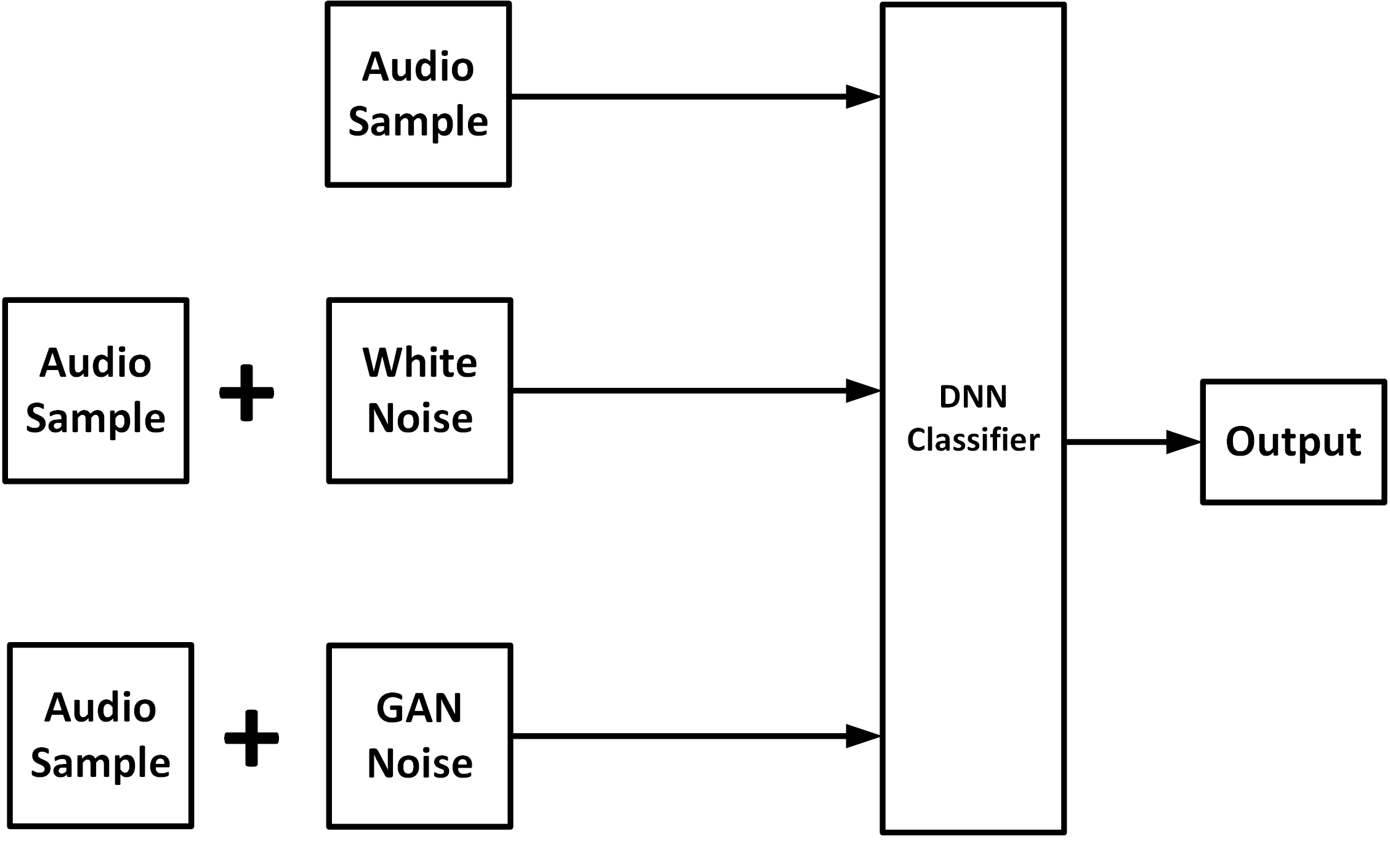}
	\caption{Solution Architecture}
	\label{fig:architecture}
\end{figure}

The audio samples are all processed into 16bit, 48kHz .wav format before being converted into spectrogram images. After the audio samples are pre-processed into spectrogram images, spectral feature extraction is carried out and the input is then fed into the Deep Neural Network (DNN) classifiers. Fig. \ref{fig:architecture} shows a diagrammatic representation of the solution architecture.

\subsection{Noise Generation Methodology}
Our solution is based on the premise that GAN generated noise, when combined with audio recordings from the smart home device, reduces the effectiveness of machine learning based inference from the audio recordings. This enhances smart home user privacy from various forms of adversaries with varying capabilities discussed in the preceding paragraph. Our results from section \ref{results} highlight more details on this. 

The GAN noise signal is generated by an external device that is permanently in the smart home user's environment and constantly producing noise signals which when combined with audio recordings from the smart home device, prevents an adversary from inferring sensitive information from the audio recordings. The noise amplitude of the external noise generator is audible for human perception but it should not exceed the acceptable noise threshold for human comfort.

In this study, we evaluate the effectiveness of the GAN noise with white noise. The white noise is generated with a python script using the same hardware for generating the GAN noise. In our evaluation, both noise samples are produced at the same amplitude to ensure consistency in the results.

\subsection{MaskGAN Overview}\label{maskgan}

In the original GAN setup introduced by \cite{goodfellow2014generative}, GANs were used to generate synthetic data samples by taking as input,  statistically independent noise samples. To the best of our knowledge, GANs have not been used to generate random noise signals. We choose to implement GANs in our approach to create audio samples as against other generative models such as Variable Autoencoders (VAE) because GANs do not introduce any deterministic bias and work better with discrete latent variables \cite{goodfellow2014generative}.

Our solution which we refer to as MaskGAN is an adaptation of Deep Convolutional Generative Adversarial Networks (DCGAN) \cite{DBLP:journals/corr/RadfordMC15}. DCGANs are a notable architecture for adversarial image generation in which a transposed convolution operation is implemented for creating high-resolution images from low-resolution feature maps. Since DCGAN ouputs 64x64 pixel images, we add two additional layers to produce two seconds of audio at 16KHz. Furthermore, the 2-dimensional convolutions are flattened into 1-dimensional with the stride factors increased twofold.

\begin{figure}[ht!]
	\centering
	\includegraphics[width=0.9\linewidth,keepaspectratio=true]{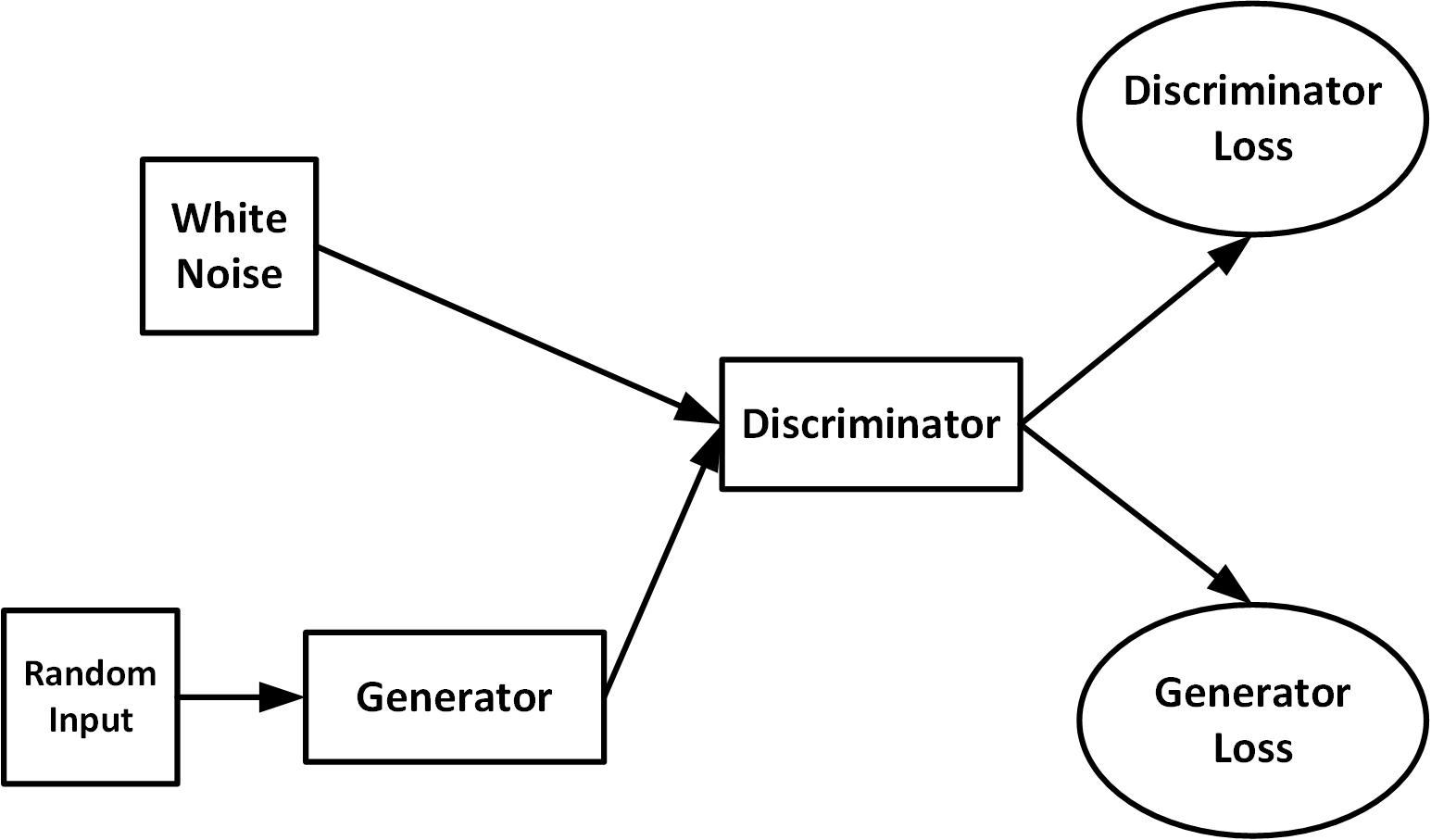}
	\caption{MaskGAN Structure}
	\label{fig:gan}
\end{figure}

Our proposed MaskGAN structure consists of two models as shown in Fig \ref{fig:gan}. The first model known as the generator tries to generate new and synthetic audio samples that are identical to the target white noise audio sample. The second model known as the discriminator performs an adversary role by trying to detect if the synthetic audio sample is real or fake, hence helping to improve the knowledge of the generator until the generator eventually succeeds in creating some synthetic audio sample which is realistic and as indistinguishable from the actual white noise audio sample as possible.

The generator model is represented as \(G(z,\theta_g)\) while the discriminator model is represented as \(D(x,\theta_d)\) where \(x\) represents the input audio samples and \(z\) represents the generated synthetic samples. The weights of the neural network also known as parameters are represented as \(\theta\). The parameters of the generator \(\theta_g\) are updated to maximize the probability that the synthetic audio is classified as the real audio dataset. The loss function of the generator network seeks to maximize \(D(G(z))\). With regards to the discriminator, the parameters are optimized to maximize the probability that the synthetic noise audio samples are classified as real audio samples. Hence, the loss function of the discriminator, seeks to maximize the function \(D(x)\) while minimizing the function \(D(G(z))\).

The minmax game between the generator and the discriminator is represented as a value function \(V(G,D)\), whereby the generator seeks to maximize the probability that its output is classified as real. In contrast, the objective of the discriminator is to minimize this probability.

The input to the MaskGAN model is a random seed and the target output is a white noise signal that has been generated from our python code. After several iterations, the generative model finally arrives at a synthetic noise signal that is indistinguishable from the digital white noise signal, based on the assessment of the discriminative model. 



\subsection{Dataset, Developmental Tools, Hardware and Software}\label{tools}
For our experimental work, we used a standalone desktop PC running windows 10 Education OS. The hardware components consist of an AMD Ryzen 7 2700X processor at 3.70GHz with 32GB of RAM and 1TB SSD storage. The graphics card is a standalone GPU - NVIDIA GeForce RTX 2080 TI with 11GB RAM. 

All software development was carried out using publicly available and open-source tools. The software code was written in Python programming language using the Spyder Integrated Development Environment (IDE), which is part of the "Anaconda software distribution". For the deep learning framework, we used the Google TensorFlow v2 deep learning framework. 

The three datasets we used represent the three inference attack case scenarios that were explored in this study, namely music genre inference (MGI), user demographics inference (UDI), and speech emotion inference (SEI). The datasets used are publicly available, and details of each dataset are further discussed in section \ref{bia_results}.

\section{Experimental Approach}
Assuming the smart home has devices that are equipped with always-listening capabilities. As discussed in section \ref{threat} above, these devices could be harnessed by an adversary to leak sensitive information from the occupants of the home. Our experiments seek to deter such leakage inference attacks using truly random noise generated by a GAN neural network model which we term as MaskGAN. The first subsection describes our approach for generating audio noise with increased randomness using our GAN solution. The next subsection \ref{random_measure} describes our approach for measuring the randomness of the GAN generated noise and performing a comparison to the white noise using two different runs tests methods. In the 3rd subsection \ref{infer_baseline}, we describe how we perform the inference attacks for three different scenarios using three different datasets and for each dataset, three different neural network models are utilized. In this step, the original audio dataset is used without adding any form of noise mitigation. In the next subsection \ref{mitigate}, we discuss our approach to mitigate the leakage of sensitive information via inference attacks with the use of the noise generated by the GAN and white noise.  In the final subsection \ref{eval} we discuss the different metrics we utilize for evaluating our methodology and results.

Since this paper focuses on information leakage from smart homes rather than eavesdropping, our case study scenarios and dataset selection best reflect this context. For example, rather than selecting datasets for automatic speech recognition such as \cite{speechcommandsv2}, we instead select datasets in which information inference is sought from the audio samples. In our "semantic preservation factor" evaluation metrics in section \ref{spf_eval}, we discuss our approach to experientially highlight the difference between both contexts and report our results in section \ref{sra_results}.  For our case study, we consider the possibility of an adversary seeking to infer what genre of music the occupants of a home prefers to listen to and therefore provide targeted ads to the user. The second case scenario demonstrates an adversary who infers the user demographics such as race and gender of the home occupants, while the 3rd case scenario discusses an adversary who seeks to infer the emotion of the home occupants. The adversary achieves this sensitive information leakage or inference attacks using machine learning or deep learning techniques applied to the audio recordings. Other possible adversary scenarios may include the possibility to allow speech recognition while blocking out contextual information leakage.

\subsection{Generate Noise Signals with GAN}\label{noise_gen}
The first step in our experimental approach entails using the GAN structure described in section \ref{maskgan} to create noise samples using white noise as the target output. As illustrated in Fig. \ref{fig:gan}, the generative model produces audio samples from a random seed and learns to improve as the discriminator determines how close the audio sample is to the white noise signal. The amplitude of the generated GAN noise does not exceed 45db in order to remain with the human comfort level as specified in \cite{fincher2009standards}

\subsection{Measuring the degree of Randomness in Noise Signals}\label{random_measure}
In the second step of our experimental approach, we compute the degree of randomness of the original sample, the white noise and the GAN noise. In this section, we use two different non-parametric approaches in determining the degree of randomness of the audio samples. 

Each audio sample is represented as a matrix of integers, with the shape representing the dimensions. For the scope of our study, we focus on notable runs tests in which upward and downward run counts are carried out for a sequence of variables, by floating the integers of the audio samples represented as an integer matrix. 

Two measures of randomness namely the Wald-wolfowitz runs test \cite{wald1940test} and the Cox-stuart test \cite{cox1955some} are used to measure and compare the degree of randomness between the three audio signals. The results are reported in sections \ref{wald} and \ref{cox}.

\subsubsection{Wald-wolfowitz Runs tests}\label{wald} The wald-wolfowitz runs tests \cite{wald1940test} considers each integer in the integer matrix representation of the audio sample as \(n\) observations with a median value. A measure of the expected runs \(E(R)=\frac{{2n_1}{n_2}}{n}+1\) and the variance \(V(R)=\frac{{2n_1}{n_2}({2n_1}{n_2}-n)}{n^2(n-1)}\) are computed respectively below to establish the statistical ratio. The equation below from \cite{wald1940test}

\begin{equation}
Z_R= \frac{(R- E(R))}{V(R)}
\end{equation}

which represents the number of runs in the representation of the audio file corresponding to its size.

\subsubsection{Cox-stuart Test}\label{cox} The Cox-Stuart test \cite{cox1955some} focuses on randomness based on negative or positive tests in data. Taking into consideration the sum of positive signs for an integer matrix representing each audio sample, a p-value is taken as a cumulative probability function for a binomial distribution of the dataset. The integer matrix representation of the dataset is grouped into pairs with the sign computed. The sign test in the equation below is used to determine if there is a trend in randomness as observed in the integer matrix representation of the audio sample. The equation below from \cite{cox1955some}
\begin{equation}
sign(x_i,x_i+c) =\left\{
                \begin{array}{ll}
                  + & if X_i = X_{i+c}\\
                  0 & if X_i \leq X_{i+c}\\
                  - & if X_i \geq X_{i+c}\\
                \end{array}\\
              \right.
\end{equation}

Thus, the p-value with a count of the positive comparisons forms the statistical ratio for the degree of randomness.

\begin{figure}[h!]
	\centering
	\includegraphics[width=0.9\linewidth,keepaspectratio=true]{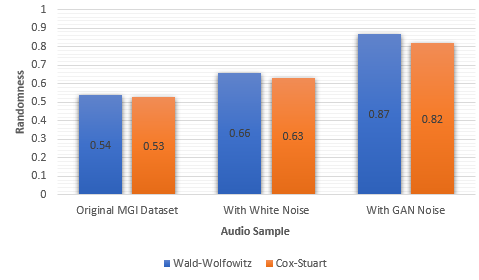}
	\caption{Randomness tests with MGI dataset}
	\label{fig:mgi_randomness}
\end{figure}

\begin{figure}[h!]
	\centering
	\includegraphics[width=0.9\linewidth,keepaspectratio=true]{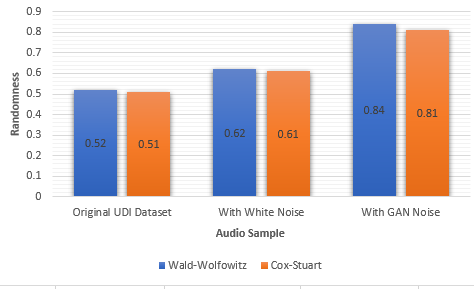}
	\caption{Randomness tests with UDI dataset}
	\label{fig:udi_randomness}
\end{figure}

\begin{figure}[h!]
	\centering
	\includegraphics[width=0.9\linewidth,keepaspectratio=true]{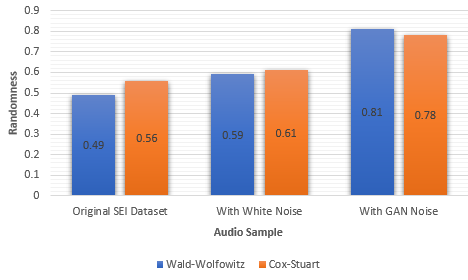}
	\caption{Randomness tests with SEI dataset}
	\label{fig:sei_randomness}
\end{figure}

\textbf{Outcome of the Runs Test for Randomness:}
For each runs test, we compute the average across the entire dataset for each inference attack case scenario mentioned in section \ref{tools}. We repeat the process of the runs test computation for each of the datasets with the white noise added and also with the GAN noise added. First, we compute the degree of randomness in the  original audio sample. We then compare the degree of randomness with  the audio sample superimposed with the white noise sample as well as the audio sample superimposed with the GAN noise. Figures \ref{fig:mgi_randomness}, \ref{fig:udi_randomness} and \ref{fig:sei_randomness} show the results of the randomness tests. The results show that on a scale of 0-1, the audio sample overlaid with the GAN noise shows more randomness based on both runs test compared to the original dataset as well as the dataset with the white noise.

\subsection{Perform Inference Attacks on Original Audio Samples}\label{infer_baseline}
Machine learning-based audio profiling of voice recordings from always listening devices can be used to infer sensitive information from a smart home user 's environment. We experiment with a total of three publicly available datasets,  to explore three types of inference attacks to leak out sensitive information about the occupants of a home. 

\subsubsection{Inferring User Music Listening Preferences from Smart Listening Devices}
We use the Free Music Archive Dataset \cite{defferrard2016fma} which is an open and easily accessible dataset suitable for evaluating several tasks in music information retrieval (MIR).  It consists of full-length and high-quality audio which includes metadata, tags. The dataset consists of 106,574 tracks from 16,341 artists and 14,854 albums, arranged in a hierarchical taxonomy of 161 genres.

\subsubsection{Inferring User Demographics from Smart Listening Devices} This task involves inferring three basic user demographics context from audio files, namely age, gender and race. The dataset used is the Mozilla common voice dataset \cite{ardila2019common}, which consists of about 51,000 voice recording samples. We use all three DNN architectures to perform multi-class, multi-label classification.

\subsubsection{Inferring Emotional Content from Smart Listening Devices}
In the third step of our experimental approach, we explore the feasibility of an adversary to infer emotional context from a user's private conversations. As earlier discussed, monetary motives such as targeted advertisements may be a factor for such an adversary in implementing this form of inference attack. For this case scenario, the Ryerson Audio-Visual Database of Emotional Speech and
Song (RAVDESS) \cite{livingstone2018ryerson} is used.

\subsection{Mitigate Sound Inference Attacks}\label{mitigate}
Our methodology entails the superimposition of the original audio samples with some form of external audio noise to prevent an adversary from inferring unwarranted sensitive information from audio recordings. The external noise is generated in with two methods. For the first method, white noise is generated. The second method entails the use of a Generative Adversarial Network (GAN) architecture, which forms the basis of our proposed solution.

\subsection{Evaluation}\label{eval}
We evaluate the effectiveness of our solution based on three metrics. The mitigated inference accuracy (MIA), the semantic preservation factor (SPF), and randomness to mitigation relationship (RTMR). First, we establish a benchmark assessment which we report in section \ref{bia_results} as the Baseline Inference Accuracy (BIA). We then proceed to evaluate our proposed solution based on the metrics described below.

\subsubsection{Mitigated Inference Accuracy}\label{mia_eval} The mitigated inference accuracy (MIA) denotes the prediction accuracy of the DNN model in inferring sensitive information from an audio dataset when the sound masking noise has been applied. We report this metric for all inference scenarios using the three different DNN architectures described in the study.

\subsubsection{Semantic Preservation Factor} \label{spf_eval} The semantic preservation factor (SPF) represents the attribute of the sound masking signal to preserve the semantics of the audio content. We use a different dataset for this experimental setup with the three DNN models to compare the SPF of both white noise and the GAN noise.

\subsubsection{Randomness to Mitigation Relationship}\label{rtmr_eval} The third evaluation metric compares the randomness in the GAN noise and white noise with the mitigation inference accuracy. For both the white noise and the GAN noise, we calculate the element of the randomness in the audio dataset when each noise sample is added, compared to the effect of the inference mitigation that was achieved. 

\section{Results}\label{results}
In this section, we report our experimental findings based on our three evaluation criteria discussed in section \ref{eval} .  In the first subsection \ref{bia_results}, we establish the effectiveness of the inference attack on all three datasets. In the next results section \ref{mia_results}, we compare the effect of both the GAN noise as well as the white noise in mitigating inference attacks for all three case scenarios. In the next results section \ref{sra_results}, we show results which demonstrate that the GAN noise is more effective in preserving the semantics of the audio compared to the white noise. In the final results section \ref{rtmr_results}, we show how the randomness for both the white noise and the GAN noise correlates with the mitigated inference accuracy for all three case scenarios.

\subsection{Baseline Inference Accuracy} \label{bia_results} 
\begin{figure}[ht!]
	\centering
	\includegraphics[width=0.9\linewidth,keepaspectratio=true]{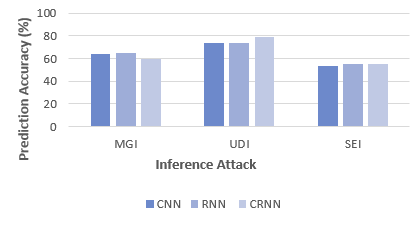}
	\caption{Baseline Inference Accuracy}
	\label{fig:bia}
\end{figure}

In the first experiment, we conduct a baseline assessment of the inference attacks against all 3 datasets for the 3 case scenarios we considered. All 3 machine learning techniques were effective in inferring information from the audio dataset with a highest achievable inference of 82\% from the CRNN model for the user demographic inference (UDI). as shown in Fig. \ref{fig:bia}. The figures reported are the best results achieved based on K-fold cross validation which was used to determine the optimal parameter settings of the neural network models.

\subsubsection{Music Genre Inference (MGI)}
As part of privacy considerations and in the context of user privacy, a user's preferences for music listening may be chosen not to be shared with external parties without their consent. An adversary may however want to infer this information for example for monetary purposes such as targeted advertisement without the user's consent.  The ability of an adversary to infer this information is demonstrated using  the Free Music Archive dataset \cite{defferrard2016fma}. We confirm using three different Deep neural network architectures that the music genre can be correctly inferred with an accuracy of up to 67\%.

\subsubsection{User Demographics Inference (UDI)}
We explore the feasibility of an adversary to infer user demographic data such as age, accent and gender from audio dataset using the Mozilla Common Voice dataset \cite{ardila2019common}. The dataset consists of about 51,000 voice recording samples of humans in 18 different languages. Our DNN models identify the demographic qualities of the speaker with an accuracy of 74\%, 71\% and 89\% respectively. When all demographic properties are combined, an accuracy of 82\% is achieved.

\subsubsection{Speech Emotion Inference (SEI)}
In our third privacy inference case scenario, we examine the ability of an adversary to infer the emotion of users from a given dataset. We use the Ryerson Audio-Visual Database of Emotional Speech and Song (RAVDESS) \cite{livingstone2018ryerson} which consists 7356 audio samples of 12 female and 12 male professional actors. Each of the actors is tasked with speaking out two lexically matched statements using a neutral North American accent.
The dataset is labeled to distinguish a total of 7 different
emotions including calm, happy, sad, angry, fearful, surprise,
and disgust expressions.

\subsection{Mitigated Inference Accuracy}\label{mia_results}

In this section, we test the privacy preservation hypothesis of the GAN noise capability in preventing sensitive information leakage in smart homes. We seek to determine if the GAN noise is more effective than white noise in preserving the privacy of smart home devices. Our results show that the noise generated by GAN results in over 45\% reduction in sensitive information leakage from smart home devices while maintaining the semantics of the audio.

\subsubsection{Mitigated Inference Accuracy (White Noise)} 
\begin{figure}[ht!]
	\centering
	\includegraphics[width=0.9\linewidth,keepaspectratio=true]{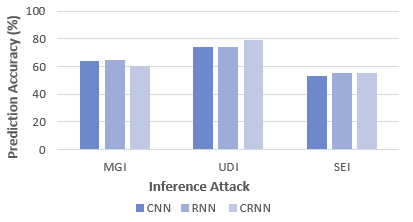}
	\caption{Mitigated Inference Accuracy (White Noise)}
	\label{fig:mia_white}
\end{figure}

In our next experiment, we tested the ability of the DNN to correctly infer sensitive information from the dataset for the case scenarios discussed above. We notice very little difference in the ability of the white noise when combined with the original audio to mitigate against information leakage. When compared to the BIA results in section \ref{bia_results}, the maximum decrease in inference that was observed when the white noise was added was less than 11\% as shown in Fig. \ref{fig:mia_white}.

\subsubsection{Mitigated Inference Accuracy (GAN Noise)}

\begin{figure}[ht!]
	\centering
	\includegraphics[width=0.9\linewidth,keepaspectratio=true]{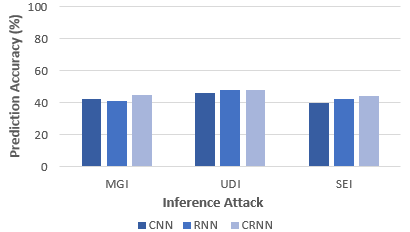}
	\caption{Mitigated Inference Accuracy(GAN Noise)}
	\label{fig:mia_gan}
\end{figure}

In our third experiment, we combine the GAN generated noise with the original audio and repeat the inference attack using the 3 DNN models. When compared to the BIA results in section \ref{bia_results}, we observe up to a maximum of 45\% decrease in inference accuracy when the GAN generated noise is added to the original audio sample as shown in Fig. \ref{fig:mia_gan}. 

\subsection{Semantic Preservation Factor} \label{sra_results}
\begin{figure}[h!]
	\centering
	\includegraphics[width=0.9\linewidth,keepaspectratio=true]{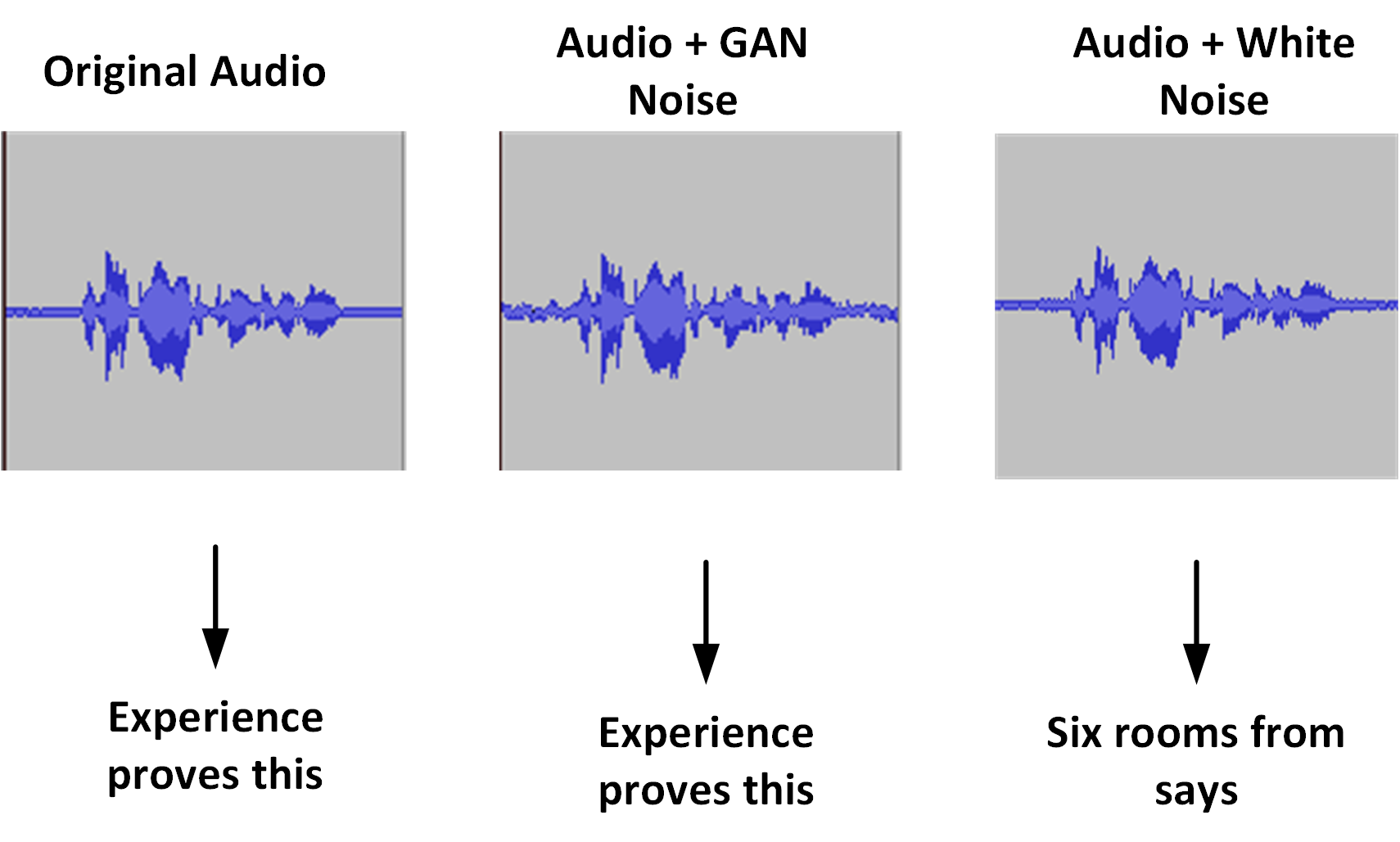}
	\caption{Illustration - Semantic Preservation Factor}
	\label{fig:spf}
\end{figure}

\begin{figure}[h!]
	\centering
	\includegraphics[width=0.9\linewidth,keepaspectratio=true]{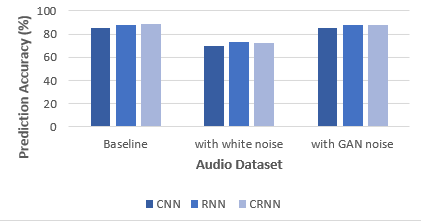}
	\caption{Semantic Preservation Factor}
	\label{fig:sra}
\end{figure}

In our fourth experiment, we demonstrate the ability of the GAN noise to preserve the semantics of the audio while effectively mitigating information leakage attacks. A visual representation of the results is show in Fig. \ref{fig:spf}
We performed speech recognition classification using the google speech commands dataset \cite{speechcommandsv2}. This fourth dataset was selected since the dataset was collected and labeled for recognizing the content of the speech. Unlike the other 3 datasets which were used in the inference attack discussed in section \ref{infer_baseline}, this dataset is more appropriate for deducing the content of the speech. The other 3 datasets were not used for semantic experiment since they were not collected and labelled for speech classification tasks.
The result of the experiment shows that the GAN noise has less impact on the DNN speech recognition classifier compared to the white noise as shown in Fig. \ref{fig:sra}. Hence, we confirm that the GAN noise does indeed preserve the utility of the device by deterring inference attack yet, maintaining the semantics of the conversation. 

\subsection{Randomness to Mitigation Relationship}\label{rtmr_results}
\begin{figure}[h!]
	\centering
	\includegraphics[width=0.9\linewidth,keepaspectratio=true]{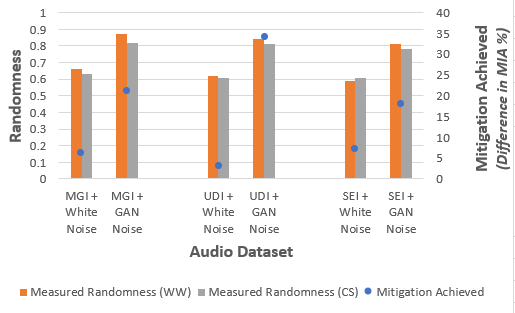}
	\caption{Randomness to Mitigation Relationship}
	\label{fig:rtmr_graph}
\end{figure}

We evaluated the relationship between randomness and the mitigation inference accuracy (MIA) in each inference attack case scenario. The result as illustrated in Fig. \ref{fig:rtmr_graph} shows that the higher the  degree of randomness, the higher the mitigation effect that the noise exhibits in deterring the inference attack. The mitigation achieved is calculated as the difference in the MIA for each case scenario with the white noise as well as the GAN noise. The GAN noise, having more randomness compared to the white noise is proven to have a higher mitigation effect.

\section{Discussion}
\subsection{White Noise and Randomness}
White noise is known to exhibit statistical characteristics that are similar to randomly generated numbers. To be considered as truly random, we expect the entity to be in fact unpredictable; but the possibility of a white noise generator to exhibit true randomness in the sense of unpredictability is questionable. Speicher et al. \cite{speicher1990new} noted that patterns can be noted in pseudo random generated white noise, based on the fact that is contains possibly predictable elements for example, a linear congruential random generator, which is typical algorithm used for producing digital noise output. Tzeng et al. \cite{tzeng2008parallel} argued that it is best to treat randomness as a property of the process that generates the signal of the white noise, not of the white noise itself. 

Hence, we establish that white noise with its pseudorandom property is thus limited in the ability as an effective measure in privacy preservation for use in audio masking for preventing sensitive information leakage attacks.

The ability of generative adversarial networks to create high dimensional data has been researched \cite{de2018pseudo} and the relationship of this high dimensionality to randomness is an object of interest. As the generator model in the GAN continuously learns to produce data samples which the adversary (discriminator) cannot predict, the randomness element in its output improves, as demonstrated in our results. Our solution entails the use of a generative model that has learnt to produce realistic noise samples of a given dataset from low-dimensional, random latent vectors.

Several recent efforts have been made to generate sound using Generative Adversarial Networks including the use of CycleGAN by \cite{asakura2019emotional}. Their approach augments an existing audio sample with emotions and can also convert speeches between emotional variations e.g. convert an angry speech into a sad speech.

We differentiate our work from other studies such as \cite{li2020practical} which use adversarial attacks to mitigate speech recognition i.e. the use of machine learning systems to determine the identity of the speaker. In this specific study, a state of the art deep neural network (DNN) known as X-vector was tested. By adding a carefully crafted inconspicuous noise to the original audio, their attack method was successful in fooling the DNN into making false predictions. The solution goes further to incorporate room impulse response (RIR) estimates while training the adversarial examples to demonstrate the effectiveness for both digital attacks as well as over-the-air attacks.



\subsection{Mitigating Privacy Inference Leakage in Digital Space Vs. Physical Space}
Sound masking in the context of this study occurs in the physical space and is more practical oriented. Factors such as the room impulse ratio is considered in deploying real and tangible audio signals to mitigate unwanted sensitive information leakage due to machine learning inference.
Traditionally, there have been several ways of attacking speech recognition systems. Adversarial examples, for instance the work of Carlini and Wagner \cite{carlini2018audio} could impact a speech recognition system to misclassify by adding a carefully crafted perturbation. This adversarial attack method was tested against speech recognition only, but not tested against speech inference. Furthermore, their attack was proven to be effective in the digital space. Similar studies \cite{alzantot2018did} \cite{carlini2018audio} have proposed solutions mostly against automatic speaker recognition in the digital space. In our threat model, we considered various adversaries, including the manufacturer or solution developer who controls the digital space and therefore, implementing a solution within the digital space will be ineffective against such adversaries. 


Also, the work of Fuxun Yu et al \cite{yu2019masker} introduced "MASKER", a solution which introduces human imperceptible adversarial perturbation into real time audio signals with significant increase in the word error rate (WER). Their work focused on mobile platforms and was not tested to work against digital voice assistants or smart home environments. Also, the solution is primarily effective only in the digital space and was not tested as an over-the-air solution.

Our method focuses on a solution that is implemented within the physical space. Since the user can perceive the sound generated, we ensure that this sound is within the audible comfort zone for human hearing of less than 45db \cite{fincher2009standards}. We tested with various amplitudes of audio signal and determined that as the amplitude increases, the effectiveness also increases. However, compared to the original audio as well as the white noise, our GAN generated noise achieves better mitigation based on several metrics.

We show that speech recognition and sound-based inference have varying and different and unique characteristics and adversarial noise techniques should be considered differently. Refer to Fig. \ref{fig:differen} where we illustrate the difference between eavesdropping and inference attacks.

\section*{Conclusion}
We proposed a novel method for mitigating sensitive information leakage in smart home environments. We highlighted a threat model whereby an adversary deploys a machine learning based inference attack on connected, always listening devices such as smart phones or smart speakers. 

Our solution is based on a generative deep learning model known as the Generative Adversarial Networks (GAN). We established from our experiments that GAN based audio samples have increased randomness compared to white noise. Also, when used for sound masking purposes, GAN generated noise can effectively mitigate machine learning based inference attacks in smart home environments while preserving the semantics of the audio conversation. 

\section*{acknowledgement}
This work was supported by the Natural Sciences and Engineering Research Council of Canada (NSERC) through the NSERC Discovery Grant program.

\bibliographystyle{ieeetr}
\bibliography{main}

\end{document}